\documentclass[10pt,twocolumn,letterpaper]{article}

\usepackage{iccv}              

\usepackage{multirow} 
\usepackage{booktabs}
\usepackage{arydshln}

\definecolor{iccvblue}{rgb}{0.21,0.49,0.74}
\usepackage[pagebackref,breaklinks,colorlinks,allcolors=iccvblue]{hyperref}

\usepackage[most]{tcolorbox}
\usepackage{listings}
\usepackage{caption}

\def\paperID{*****} 
\def\confName{ICCV}
\def\confYear{2025}

\title{WP-CLIP: Leveraging CLIP to Predict Wölfflin's Principles in Visual Art}

\author{
Abhijay Ghildyal \quad Li-Yun Wang \quad Feng Liu \\
Portland State University \\
{\tt\small \{abhijay, liyuwang, fliu\}@pdx.edu}
}

\begin{document}
\maketitle
\begin{abstract}

Wölfflin's five principles offer a structured approach to analyzing stylistic variations for formal analysis. However, no existing metric effectively predicts all five principles in visual art. Computationally evaluating the visual aspects of a painting requires a metric that can interpret key elements such as color, composition, and thematic choices. Recent advancements in vision-language models (VLMs) have demonstrated their ability to evaluate abstract image attributes, making them promising candidates for this task. In this work, we investigate whether CLIP, pre-trained on large-scale data, can understand and predict Wölfflin's principles. Our findings indicate that it does not inherently capture such nuanced stylistic elements. To address this, we fine-tune CLIP on annotated datasets of real art images to predict a score for each principle. We evaluate our model, WP-CLIP, on GAN-generated paintings and the Pandora-18K art dataset, demonstrating its ability to generalize across diverse artistic styles. Our results highlight the potential of VLMs for automated art analysis. Code: \url{https://github.com/abhijay9/wpclip}

\end{abstract}    
\section{Introduction}
\label{sec:intro}

In formal analysis of visual art, elements such as line, color, shape, texture, and space serve as fundamental building blocks that artists manipulate to convey ideas and emotions~\cite{arnheim1954art,gombrich1960study,hatt2006art,frank2019prebles,foka2024framework}. The arrangement of these elements within a composition not only influences aesthetic appeal but also communicates meaning beyond mere representation. Heinrich Wölfflin identified key stylistic characteristics in art and introduced five contrasting pairs of principles: linear vs. painterly, closed vs. open, planar vs. recessional, multiplicity vs. unity, and absolute vs. relative~\cite{wolfflin1950principles}. These principles provide a structured framework for analyzing visual art across different styles and periods. \textit{Originally developed to examine stylistic evolution, especially between the Renaissance and Baroque periods, Wölfflin’s framework remains widely applicable for understanding artistic composition and the formal qualities of paintings.} Quantifying these principles computationally can enable automated analysis of artistic styles, offering new insights into artistic techniques, comparisons of art movements and styles, and a deeper understanding of visual composition.

\begin{figure}[t!]
  \centering
  \includegraphics[width=0.85\columnwidth]{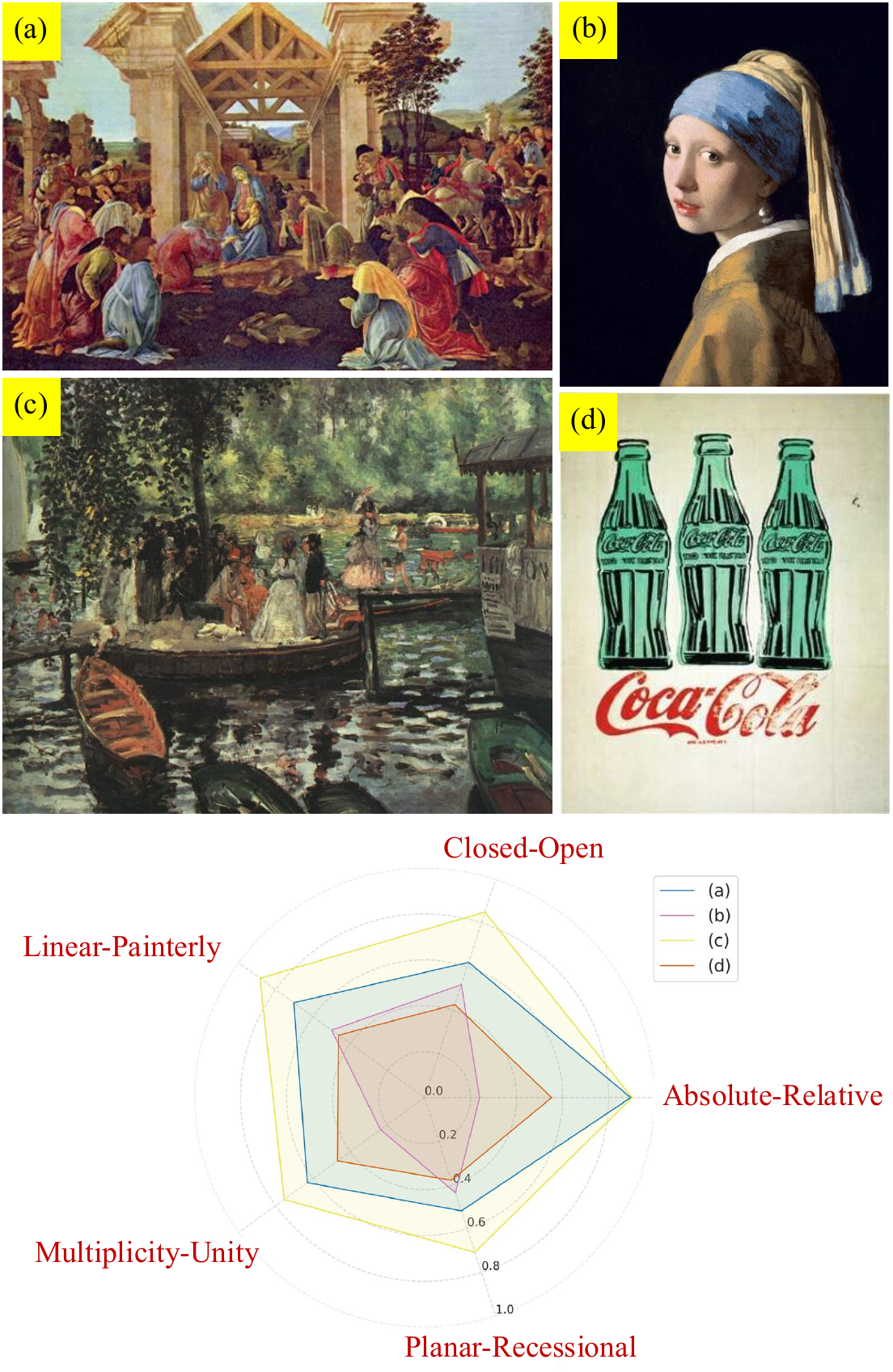}
  \caption{Examples of Wölfflin’s principles predicted for famous paintings using our metric, WP-CLIP.}
  \label{Figure:pandora_art}
  \vspace{-0.2in}
\end{figure}

\textit{Quantifying the formal analysis of paintings using visual elements is challenging due to the limited availability of direct annotations~\cite{kim2022proxy}. Moreover, predicting artistic concepts from visual elements using the five Wölfflin's principles is challenging due to their complexity and nuance.} In this work, we utilize the annotated dataset on Wölfflin's principles provided by \cite{djwaga} to develop a metric that predicts scores for each principle based on an image. Currently, no single metric exists that predicts all five of Wölfflin's principles in visual art. Recent studies have demonstrated that vision-language models (VLMs) can evaluate or score abstract attributes in images~\cite{wang2022exploring}. These VLM-based metrics are gaining significant attention, as such models are pre-trained on large and diverse datasets, enabling them to learn both semantic and visual information. This allows them to generalize effectively to tasks involving abstract image characteristics, such as style, mood, creativity, and aesthetic quality. Specifically, CLIP-IQA leverages CLIP in a zero-shot manner using the antonym pair \textit{``good photo''} and \textit{``bad photo''}, which naturally generalizes to other antonym pairs related to abstract perception, such as happy, natural, scary, complex, and more. We take a similar approach by treating the Wölfflin's pairs as antonym pairs and first investigate whether CLIP or CLIP-IQA understands Wölfflin's principles. Our findings indicate that they do not. Therefore, we further fine-tune CLIP to develop a more effective metric.

Our metric evaluates the paintings by assigning a range of scores for each of the Wölfflin's principles, as presented in Figure~\ref{Figure:pandora_art}. Renoir’s `La Grenouillère' (Figure~\ref{Figure:pandora_art}(c)) exhibits a more painterly and open composition compared to the other paintings. Additionally, it is distinctly recessional, conveying a greater sense of depth, whereas the compositions in Figure~\ref{Figure:pandora_art}(b) and Figure~\ref{Figure:pandora_art}(d) appear more planar. Vermeer’s `Girl with a Pearl Earring' (Figure~\ref{Figure:pandora_art}(b)) demonstrates higher absolute clarity and unity relative to the other paintings. Both Andy Warhol’s pop art piece (Figure~\ref{Figure:pandora_art}(d)) and Vermeer’s painting (Figure~\ref{Figure:pandora_art}(b)) exhibit a more closed composition compared to the paintings in Figure~\ref{Figure:pandora_art}(a) and Figure~\ref{Figure:pandora_art}(c). The `Adoration of the Magi' by Sandro Botticelli (Figure~\ref{Figure:pandora_art}(a)) and Renoir’s `La Grenouillère' (Figure~\ref{Figure:pandora_art}(c)) display greater unity, as the colors blend into one another. In contrast, `Girl with a Pearl Earring' (Figure~\ref{Figure:pandora_art}(b)) and Warhol’s work (Figure~\ref{Figure:pandora_art}(d)) align more with the principle of multiplicity, as the figures and objects are distinct with well-defined boundaries. More details on these principles is provided in Section~\ref{five_wp}.

In summary, our contributions are as follows:
\begin{itemize}
    \item We propose WP-CLIP, the first vision-language-based metric that predicts five scores corresponding to Wölfflin's principles for an art image. To develop this metric, we fine-tune the CLIP-ViT-B/32 model on an existing dataset that contains real art images annotated with Wölfflin's principles.
    \item We evaluate the accuracy of our WP-CLIP model on GAN-generated art test images, despite being trained on real art images, demonstrating its ability to generalize across different dataset distributions. Furthermore, we compare WP-CLIP against Gemini-2.5-pro~\cite{google_gemini25pro_2025} showing that even advanced proprietary models fall short in accurately analyzing these principles.
    \item We assess WP-CLIP's effectiveness in analyzing art movements from the Pandora-18K art dataset~\cite{florea2017artistic}, highlighting its versatility in art analysis.
    \item We further analyze WP-CLIP's predicted scores using t-SNE to evaluate their effectiveness in distinguishing synthetic from real art. The same approach is used to explore whether these principles can identify photographic styles and their relationships.
    \item Finally, we use Disco Diffusion to generate art by leveraging guidance from both CLIP and WP-CLIP, enhancing the generative process and creating diverse artworks.
\end{itemize}

\section{Related Work}
\label{sec:related}

Previously studies have utilized specific features such as brushstrokes, emotion, color-based attributes, and their combination to analyze visual art~\cite{liu2021novel,zhang2022analysis,sigaki2018history,zou2014chronological}. However, Wölfflin's principles differ from these as they are more intricate, and nuanced and cannot be assessed for a painting solely based on these spatial characteristics~\cite{cetinic2020learning,djwaga}. This suggests that analyzing visual art demands a deeper and more comprehensive approach. Recent studies focus on evaluating paintings through the embedding space of neural networks, harnessing their capability to capture intricate patterns instead of depending on manually designed features~\cite{castellano2021deep,gairola2020unsupervised,bianco2019multitask,cetinic2018fine,garcia2019context,wright2022artfid,conde2021clip}. In our work, we have a similar objective of developing a metric that analyzes visual art using the embeddings of a pre-trained model. These studies focus more on classification, style similarity, style grouping, and, more recently, descriptive analysis~\cite{bin2024gallerygpt}, as well as out-of-distribution (OOD) detection using PCA on CLIP's vision embedding~\cite{khan2024ai}. In comparison, we focus on predicting Wölfflin's principles by directly utilizing the vision and language embeddings of CLIP~\cite{radford2021learning}. We aim to quantify formal analysis, inspired by recent studies~\cite{foka2024framework,tao2024does}. While these studies used LLM prompting, we developed a metric based on annotations on the pairs in Wölfflin's principles. We believe that, in the future, LLMs can assist in extending our metric to other concept pairs.

Elgammal~et.~al. fine-tuned an ImageNet-pretrained convolutional neural network (CNN) on the WikiArt dataset to classify various art styles~\cite{elgammal2018shape}. They found that visualizing the modes of variation provided valuable insights into how the networks consistently captured the evolution and characteristics of art styles. By performing dimensionality reduction using Principal Component Analysis (PCA) and Locally Linear Embedding (LLE), they observed a high correlation between the first and second dimensions of the network's intermediate activations and Wölfflin's principles. This approach allowed them to investigate how well the network captured deeper stylistic elements beyond the basic art style classification it was designed for. Cetinic et al. used a similar approach, leveraging the features learned by the model as embeddings and training on the annotated data collected by Elgammal~et.~al. to quantify and predict Wölfflin's pairs~\cite{cetinic2020learning}. In these models, separate models were trained to predict a score for each Wölfflin's principle. In contrast, we developed a single model to predict all five of Wölfflin's principles using both the image and language embeddings of the Vision-Language Model CLIP. 
\section{Method} \label{sec:method}

Extending CLIP to predict artistic concept score based on the five Wölfflin's principles is challenging due to the contrastive and ambiguous nature of these aesthetic principles. This complexity can be understood from the descriptions provided in Section~\ref{five_wp}. To successfully adapt CLIP for this task, we introduce the dataset used for CLIP finetuning in Section \ref{dataset}. We then outline our training and inference strategies for artistic style assessment in Section \ref{train_test}.

\subsection{Wölfflin's Five Principles}\label{five_wp}

The principles consist of five scales, each with two contrasting categories:
\begin{enumerate}
    \item Linear vs. Painterly: Art can be characterized by either a focus on clear, linear outlines or a more fluid, painterly approach emphasizing color and texture.
    \item Closed vs. Open: Closed forms have defined boundaries, while open forms suggest an incomplete or infinite space beyond the artwork’s edges.
    \item Absolute vs. Relative Clarity: Absolute clarity presents objects with sharp details for realism, while relative clarity contextualizes objects, using techniques like atmospheric perspective to convey depth and mood.
    \item Planar vs. Recessional: Planar art emphasizes flatness and surface, while recessional art creates depth through perspective, focusing on the illusion of space.
    \item Multiplicity vs. Unity: Multiplicity incorporates various elements or perspectives with clearly defined figures and objects, whereas unity focuses on harmony and coherence, with colors appearing seamlessly blended within the artwork.
\end{enumerate}
These contrasting categories, which make up Wölfflin's five principles, are valuable for the formal analysis of visual art styles in paintings. Please note that Wölfflin acknowledged the interrelatedness and overlap among artistic principles, and our model captures these nuances effectively. 

\subsection{Dataset used}\label{dataset}

\begin{figure}[t]
  \centering
  \includegraphics[width=0.95\columnwidth]{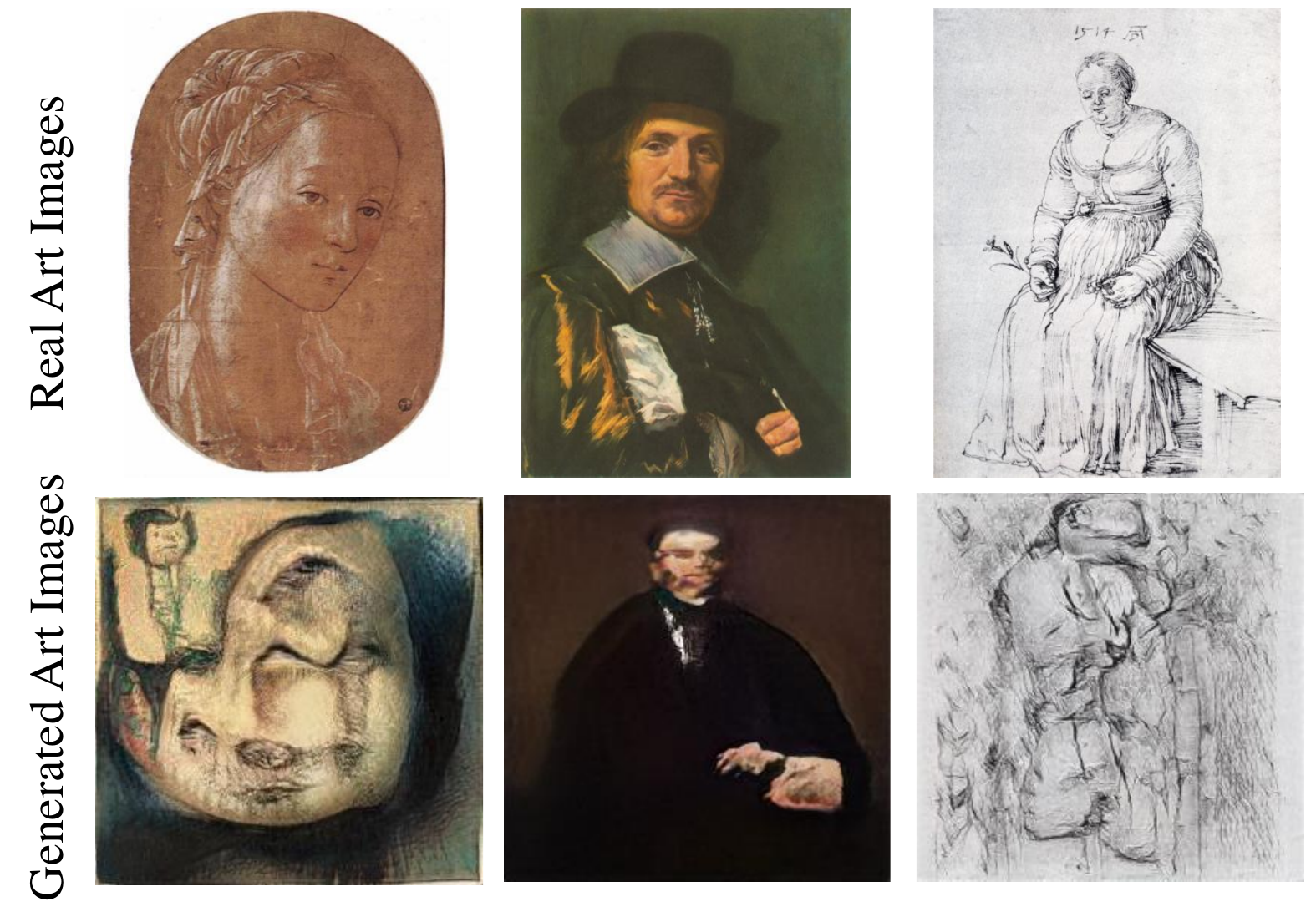}
  \caption{Samples from the Wölfflin's Affective Generative Analysis dataset. We use the 1000 real art images for training and the 800 GAN-generated art images as our test set.}
  \label{Figure:realVsGen}
  \vspace{-0.1in}
\end{figure}

In a recent study, Jha et al. annotated Wölfflin's principles on images of 1,000 real paintings~\cite{djwaga}. The selected images are diverse, with 170 artworks chosen from each century spanning from the 1400s to the 2000s. To facilitate the annotation process, they developed an interface that first trains participants to identify each principle before starting the task. We used this dataset to train our Wölfflin's principles CLIP (WP-CLIP) model. Furthermore, Jha et al. annotated Wölfflin's principles on 800 art images, with 400 generated using StyleGAN2~\cite{karras2020analyzing} and 400 using StyleCAN (Creative Adversarial Networks)~\cite{elgammal2017can}. These 800 generated images serve as our test set. Figure~\ref{Figure:realVsGen} provides examples from the training and test sets, showcasing both real and generated art images. The scores for each principle pair (e.g., Linear vs. Painterly) range from 0 to 1, indicating that the closer a score is to 1, the more Painterly the image is, and vice versa.

\subsection{Training and Inference}\label{train_test}

\noindent  \textbf{Fine-tuning CLIP to understand Wölfflin’s principles}. 
We first tested whether CLIP (ViT-B/32) and CLIP-IQA (which uses ResNet-50) can understand Wölfflin's principles. As shown in Table~\ref{table:srcc}, we found that these principles are not inherently captured in the data that CLIP was trained on. Despite being a foundation model trained on a large-scale vision and language dataset, CLIP shows no correlation with Wölfflin's pairs. Therefore, we fine-tune the CLIP (ViT-B/32) model using the formulation employed in CLIP-IQA, where CLIP (ResNet-50) was originally used.  

We utilize CLIP (ViT-B/32)'s pre-trained text and image encoders. First, we compute the similarity scores between a test image and its corresponding textual prompts using the following equation:  

\begin{equation}
s = f_I \cdot f_T
\end{equation}

where \( f_I \) is the encoded image feature, and \( f_T \) is the encoded text feature.  

For each antonym pair of texts (e.g., \( wp^{t1} = \) "Linear" and \( wp^{t2} = \) "Painterly"), we compute their respective similarity scores as  

\begin{equation}
s_{wp^{t1}} = f_I \cdot f_{wp^{t1}}, \quad s_{wp^{t2}} = f_I \cdot f_{wp^{t2}}
\end{equation}

The final score for an antonym pair is calculated as  

\begin{equation}
\label{eq:wp_score}
\hat{s}_{wp} = \frac{s_{wp^{t1}}}{s_{wp^{t1}} + s_{wp^{t2}}}
\end{equation}

Thus, each \(\hat{s}_{wp} \in [0,1]\).  

Our loss function is the mean squared error (MSE) between the predicted score and the ground truth score \( s^{gt}_{wp} \), as collected by~\cite{djwaga}:  

\begin{equation}
\mathcal{L}_{wp} = \text{MSE}(\hat{s}_{wp}, s^{gt}_{wp})
\end{equation}

Since our goal is to fine-tune our modified CLIP framework to predict a score for each of the five Wölfflin's principles, we aggregate the MSE loss across all principles. 

The total loss is computed as  

\begin{equation}
\mathcal{L}_{\text{total}} = \sum_{i=1}^{5} \mathcal{L}_{wp_i}
\end{equation}

\noindent \textbf{Inference}. After fine-tuning CLIP's text and image encoders on Wölfflin's principles, represented as antonym pair text prompts and their corresponding images, we compute the five scores for each principle using Equation~\ref{eq:wp_score}. 
\section{Experiments} \label{sec:exp}

\begin{table}[t]
\centering
\small
\begin{tabular}{rc}
\toprule
Wölfflin's Principle & MSE with GT-Score \\
\midrule
Absolute-Relative           & 0.0269                     \\ 
Closed-Open                 & 0.0178                     \\ 
Linear-Painterly            & 0.0151                     \\ 
Multiplicity-Unity          & 0.0248                     \\ 
Planar-Recessional          & 0.0182                     \\ 
\midrule
Mean               & 0.0206            \\ 
\bottomrule
\end{tabular}
\caption{Mean Squared Error (MSE) computed against ground-truth Wölfflin's principle scores (GT-Scores).}
\label{table:wolfflin_mse}
\vspace{-0.15in}
\end{table}

\subsection{Accuracy}\label{sec:accuracy}

In Table~\ref{table:wolfflin_mse}, we demonstrate that our WP-CLIP model accurately predicts the WP-Score for images in the test dataset, achieving low MSE for each principle. Although our metric is fine-tuned on 1,000 real art images, it maintains strong performance on the test dataset, which consists exclusively of generated images.

Additionally, we evaluate whether the WP-CLIP metric can accurately rank images for each principle. As shown in Table~\ref{table:srcc}, the rankings predicted by WP-CLIP exhibit a positive correlation with the ground truth. In contrast, models such as CLIP and CLIP-IQA, which are not designed for this task, perform poorly, as expected. This underscores the necessity of employing WP-CLIP, as it facilitates more accurate predictions in this context by effectively capturing the nuances and complexities of the Wölfflin's principles.

\begin{table}[t]
\centering
\small
\begin{tabular}{r c c c}
\toprule
Wölfflin's Principle & CLIP & CLIP-IQA & WP-CLIP (Ours) \\
\midrule
Absolute-Relative           & -0.06         & -0.02            & \textbf{0.54}          \\ 
Closed-Open                 & -0.04         & -0.04            & \textbf{0.39}          \\ 
Linear-Painterly            & 0.05          & -0.10            & \textbf{0.57}          \\ 
Multiplicity-Unity          & 0.06          & -0.04            & \textbf{0.30}          \\ 
Planar-Recessional          & 0.06          & 0.06             & \textbf{0.33}          \\
\bottomrule
\end{tabular}
\caption{Spearman Rank Correlation Coefficient (SRCC) for Wölfflin's Principles.}
\label{table:srcc}
\end{table}

\begin{figure}[ht!]
  \centering
\begin{tcblisting}{
  title=\textbf{Prompt for Analyzing W\"olfflin Principles},
  colback=white, 
  colframe=black,
  coltitle=white,
  fonttitle=\bfseries,
  enhanced,
  boxrule=0.8pt,
  arc=4pt,
  sharp corners=south,
  top=0pt,
  bottom=0pt,
  left=3pt,
  right=1pt,
  listing only,
  listing options={
    basicstyle=\ttfamily\footnotesize,
    keywordstyle=\color{orange}\bfseries,
    stringstyle=\color{blue},
    identifierstyle=\color{black},
    showstringspaces=false,
    tabsize=2,
    breaklines=true,
    breakatwhitespace=true,
    literate={ö}{{\"o}}1
  }
}
You are an art critic skilled in formal analysis. Using Wölfflin's five principles of art criticism, conduct a formal analysis of the two paintings shown in the figure.

Evaluate the paintings on the Left and Right according to the following principle: Linear style vs Painterly style.

Respond only with a valid JSON in the format shown below:

{
  "Left painting has more Linear style": true|false,
  "reasoning": "Brief explanation in 200 words of why you think the left painting has more Linear style and the right painting has more Painterly style, or vice versa."
}
\end{tcblisting}
  \vspace{-0.1in}
  \caption{The default prompt used for evaluating the Gemini-2.5-pro's ability to reason about Wölfflin's principles.}
  \label{fig:genai_eval}
  \vspace{-0.1in}
\end{figure}

\subsection{Comparison with Gemini-2.5-pro}

While variants of CLIP struggle with accurately predicting Wölfflin's principles scores, more advanced vision-language models (VLMs), particularly proprietary ones, may already possess the capability to perform well on this task. Currently, proprietary VLMs are among the most advanced models in terms of reasoning capabilities, consistently outperforming open-source VLMs on benchmarks involving complex mathematical reasoning~\cite{lu2024mathvista}, cross-disciplinary multimodal understanding~\cite{yue2024mmmu}, physical world comprehension~\cite{chow2025physbench}, and visual quality assurance in video games~\cite{taesiri2025videogameqa}. In our evaluation, we compare against Gemini-2.5-pro~\cite{google_gemini25pro_2025}, a leading model across these benchmarks.

We conducted an initial evaluation using popular WikiArt images and a carefully crafted prompt, as illustrated in the Figure~\ref{fig:genai_eval}. Since proprietary VLMs typically may not output a regression score like our model, hence we instead tasked Gemini-2.5-pro with comparing two paintings based on a specific Wölfflin's principle. The model accurately assessed all the initially selected pairs, demonstrating a clear understanding of the task. 

Encouraged by these results, we proceeded to test the model on pairs of GAN-generated images from the test dataset used earlier in Section~\ref{sec:accuracy}. For each Wölfflin's principle, we randomly selected 20 image pairs in which the ground-truth score difference between the images was at least 1 or 2 on a 1–5 scale, ensuring sufficient perceptual distinction within each pair. This approach resulted in two distinct sets of 100 pairs: Set 1, which is more challenging, comprises pairs with a mean GT-score difference of 1.16 and a standard deviation of 0.26; Set 2 is comparatively easier, consisting of pairs with score differences of 2 or more. As shown in Table~\ref{table:gemini}, WP-CLIP significantly outperforms Gemini-2.5-pro, achieving 71\% and 91\% accuracy on Set 1 and Set 2 respectively, compared to 57\% and 65\% by Gemini-2.5-pro. Linear versus Painterly style was the easiest principle to predict, whereas Multiplicity versus Unity was the most challenging principle for both models.

\begin{table}[t]
\centering
\small

\begin{tabular}{r c c}
\toprule
\multirow{2}{*}{Wölfflin's Principle} & \multicolumn{2}{c}{Accuracy (Count / \%)} \\
\cmidrule{2-3}
 & Gemini-2.5-pro & WP-CLIP (Ours) \\
\midrule
\multicolumn{3}{l}{\textit{Set 1: $|\text{GT}_{1} - \text{GT}_{2}| > 1$ with mean~$\pm$~std: $1.16 \pm 0.26$}}  \\
\hdashline
Absolute-Relative                    & 11 / 55\%            & \textbf{16 / 80\%}          \\
Closed-Open                          & 10 / 50\%            & \textbf{13 / 65\%}          \\
Linear-Painterly                      & \textbf{17 / 85\%}           & \textbf{17 / 85\%}          \\
Multiplicity-Unity                    & 10 / 50\%            & \textbf{11 / 55\%}          \\
Planar-Recessional                    & 9 / 45\%             & \textbf{14 / 70\%}          \\ 
Total     & 57\% & \textbf{71\%} \\
\midrule
\multicolumn{3}{l}{\textit{Set 2: $|\text{GT}_{1} - \text{GT}_{2}| > 2$ with mean~$\pm$~std: $2.09 \pm 0.14$}}  \\
\hdashline
Absolute-Relative                    & 13 / 65\%            & \textbf{18 / 90\%}          \\
Closed-Open                          & 12 / 60\%            & \textbf{19 / 95\%}          \\
Linear-Painterly                      & \textbf{18 / 90\%}           & \textbf{20 / 100\%}          \\
Multiplicity-Unity                    & 8 / 40\%            & \textbf{17 / 55\%}          \\
Planar-Recessional                    & 14 / 70 \%             & \textbf{17 / 85\%}          \\ 
Total     & 65\% & \textbf{91\%} \\
\bottomrule
\end{tabular}
\caption{Comparing WP-CLIP(ours) with Gemini-2.5-pro.}
\label{table:gemini}
\vspace{-0.1in}
\end{table}

\subsection{Analyzing Major Art Movements}

\begin{figure}[t!]
  \centering
  \includegraphics[width=0.9\columnwidth]{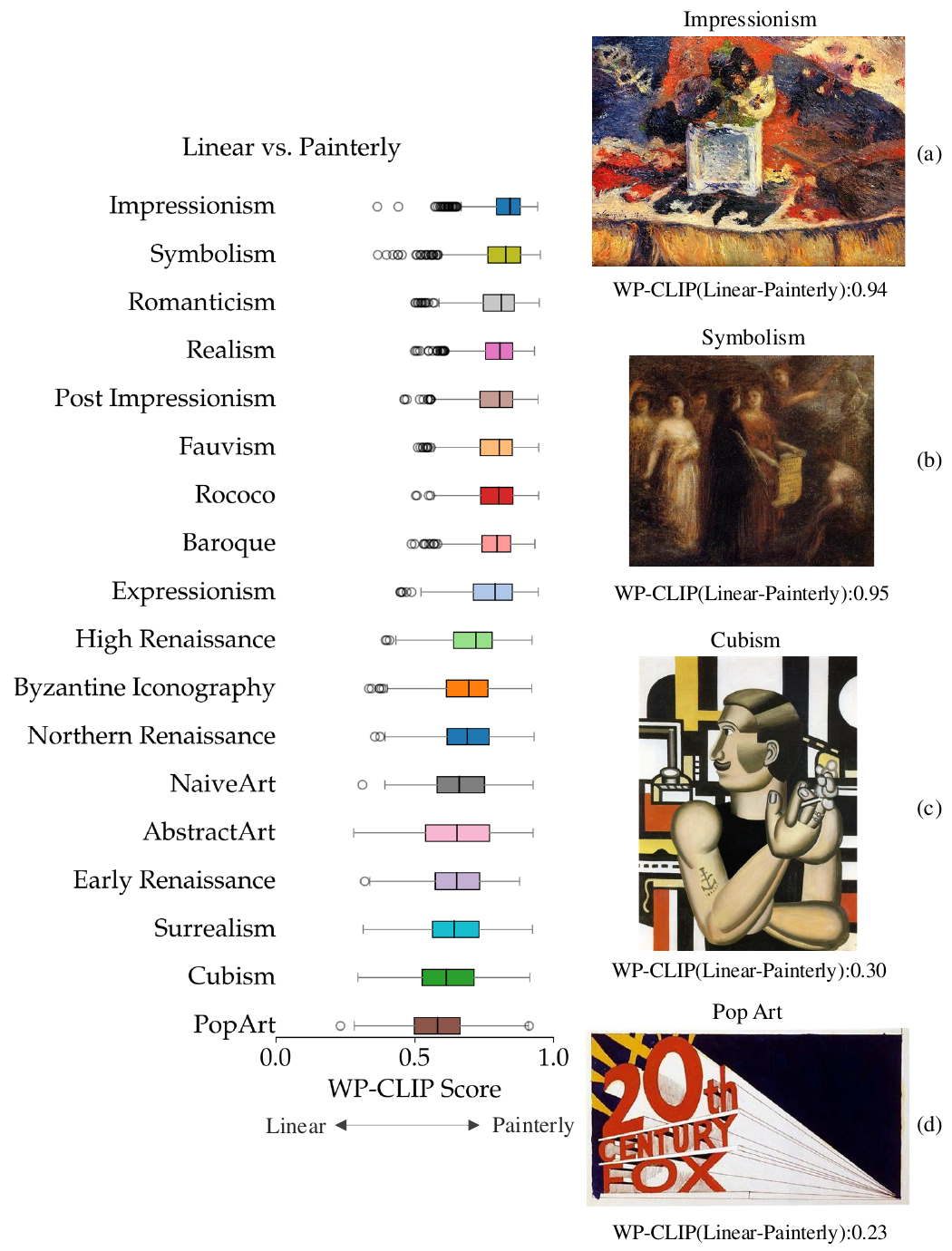}
  \caption{Predicted Linear-Painterly scores\protect\footnotemark $ $ by WP-CLIP for 18 art movements in the Pandora-18k art dataset.}
  \label{Figure:art_style_corr_lp}
  \vspace{-0.15in}
\end{figure}

Since we know that our model works well on the test dataset. In this section, we analyze the major art movements using the predicted Wölfflin's principles and sort the paintings according to predicted Wölfflin's principles. This serves as a qualitatively examination for the model, since studying art history via Wölfflin's principles has already been established.  Specifically, we evaluated our WP-CLIP on the Pandora-18k art dataset for artistic movement recognition~\cite{florea2017artistic}. This dataset was acquired in three steps. First, images were collected from various websites online and also WikiArt, along with their corresponding art movement labels. Next, non-art experts manually reviewed the images and removed those containing sculptures or 3D objects. Finally, an art expert reviewed the entire dataset and filtered out ‘non-artistic’ images. This three-step process resulted in a dataset of 18,040 artistic images spanning 18 art movements (e.g., Rococo, Symbolism, and Romanticism).


\begin{figure}[t]
  \centering
  \includegraphics[width=0.9\columnwidth]{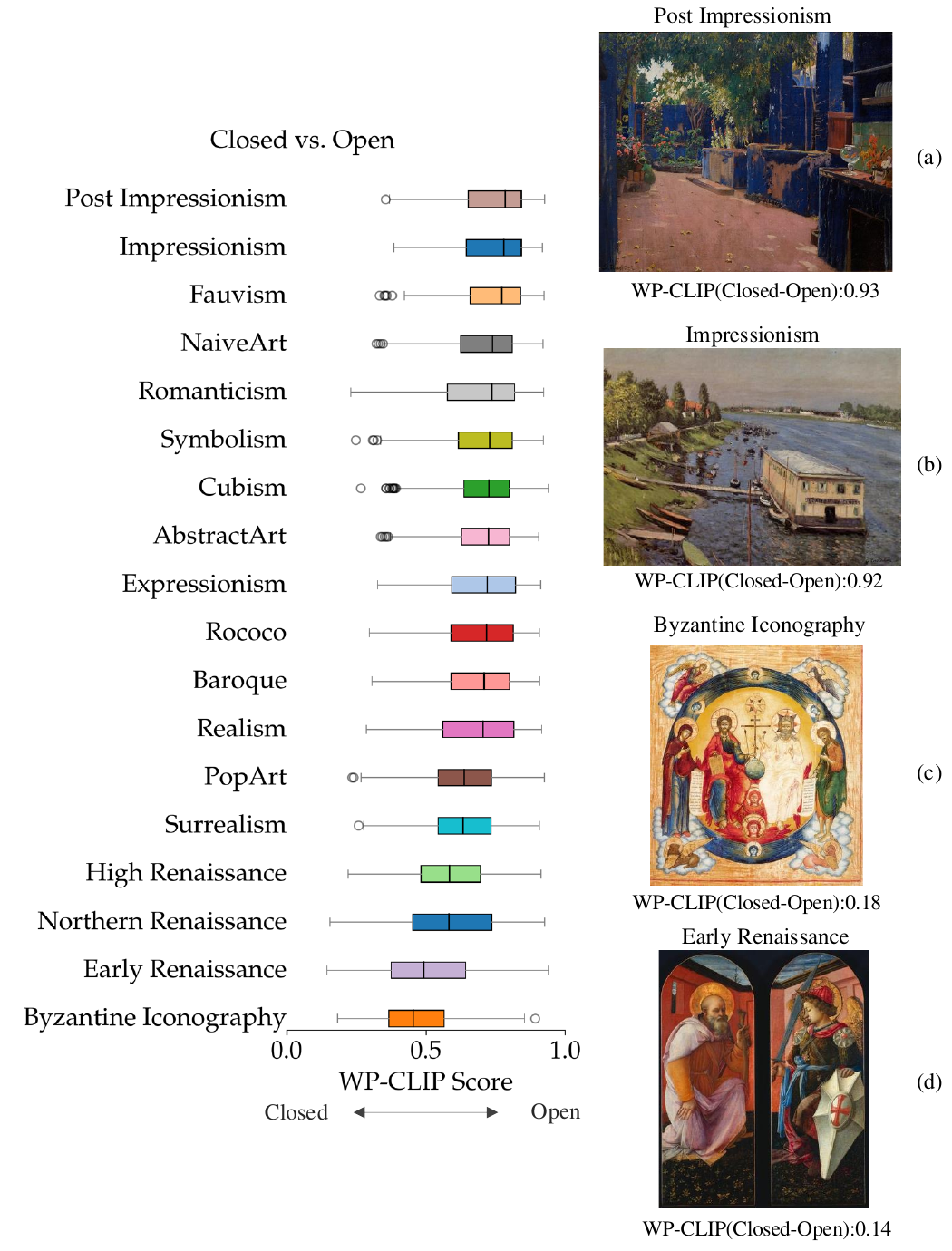}
  \caption{Predicted Closed-Open scores by WP-CLIP for 18 art movements in the Pandora-18k art dataset.}
  \vspace{-0.15in}
  \label{Figure:art_style_corr_co}
\end{figure}

According to the Linear principle, the composition emphasizes outlines and contours, with clearly defined edges that structure the artwork (e.g., Renaissance art). Conversely, according to the Painterly principle, the composition focuses on light, color, and texture, with softer and less distinct boundaries (e.g., Baroque art). As seen in Figure~\ref{Figure:art_style_corr_lp}, paintings from Impressionism, Symbolism, Romanticism, and Baroque are classified as more Painterly, while Cubism and Pop Art are categorized as more Linear. The predicted order by our metric closely aligns with the analysis by Cetinic et al. on the WikiArt dataset~\cite{cetinic2020learning}, where art movements have been sorted by the Linear versus Painterly principle. 

\footnotetext{WP-CLIP scores shown for the art images on the right (here and in later figures) refer to individual images, not the overall art movement.}

\begin{figure}[t]
  \centering
  \includegraphics[width=0.9\columnwidth]{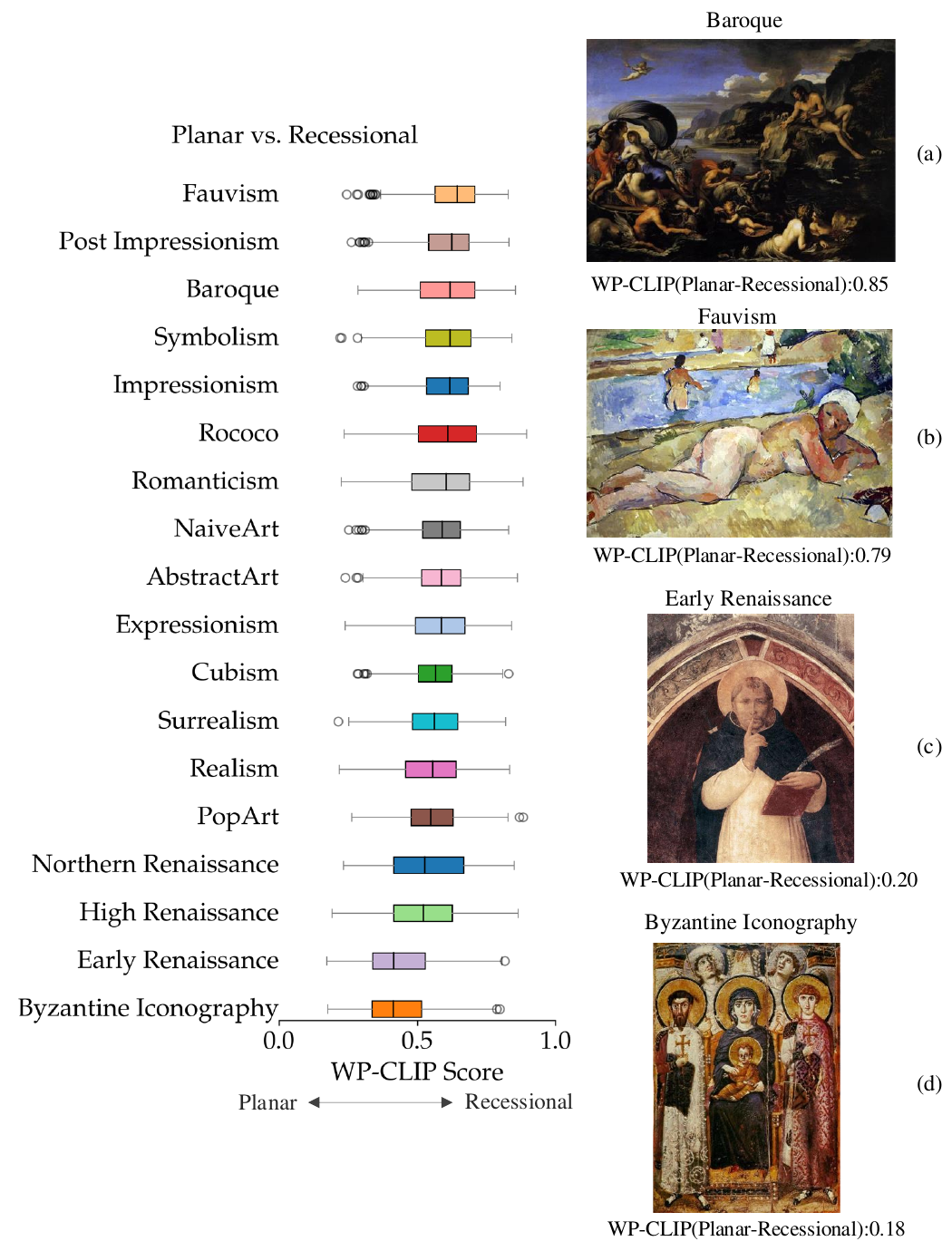}
  \caption{Predicted Planar-Recessional scores by WP-CLIP for 18 art movements in the Pandora-18k art dataset.}
  \vspace{-0.15in}
  \label{Figure:art_style_corr_pr}
\end{figure}

\begin{figure}[t]
  \centering
  \includegraphics[width=0.9\columnwidth]{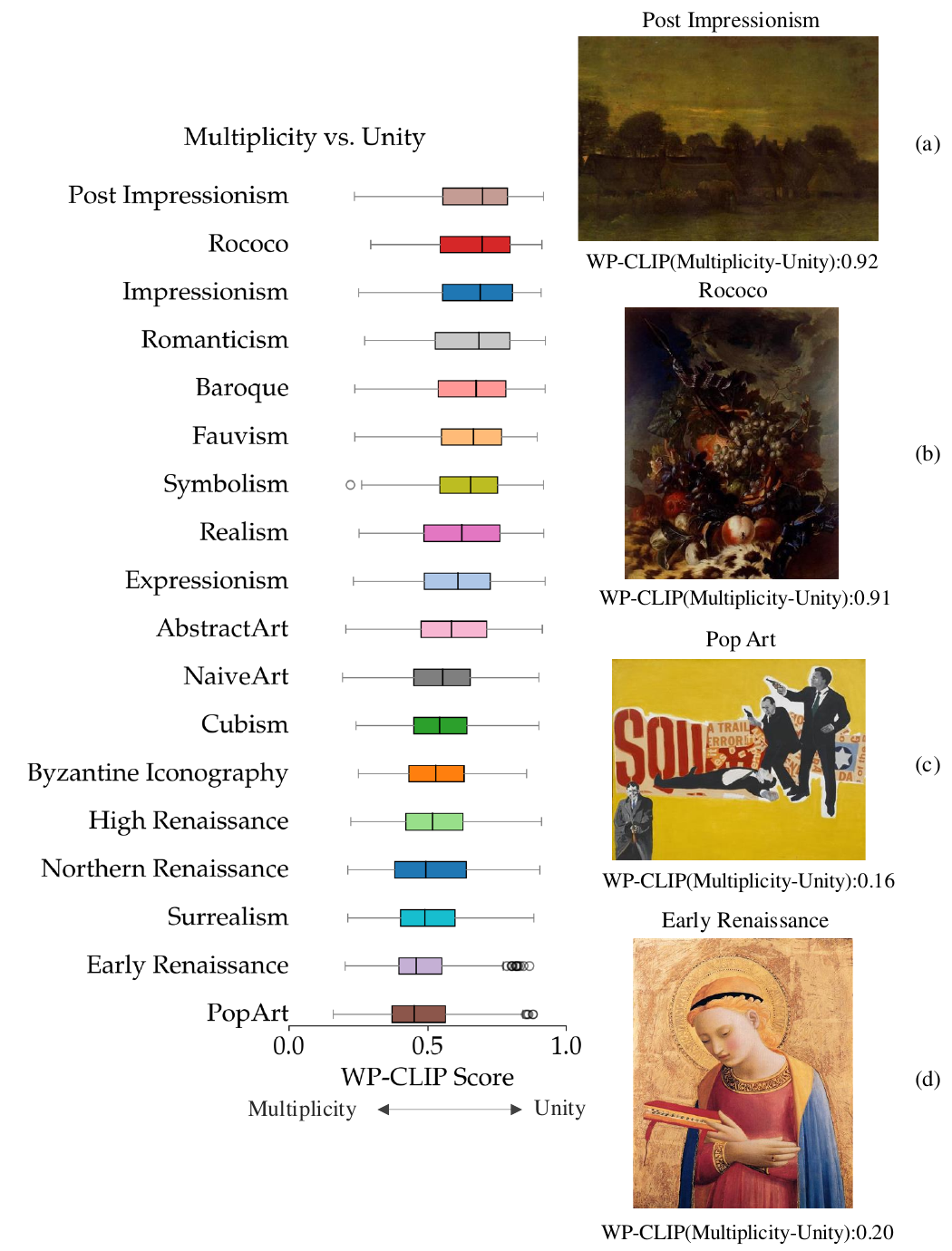}
  \caption{Predicted Multiplicity-Unity scores by WP-CLIP for 18 art movements in the Pandora-18k art dataset.}
  \vspace{-0.15in}
  \label{Figure:art_style_corr_mu}
\end{figure}

The painting `Flowers and carpet (Pansies)' by Paul Gaugin (Figure~\ref{Figure:art_style_corr_lp} (a)) is considered more Painterly by our WP-CLIP metric as Gaugin used thick brushstrokes and expressive colors. Similarly, `To Robert Schumann' by Henri Fantin-Latour (Figure~\ref{Figure:art_style_corr_lp} (b)) is also categorized as more Painterly, as the group portrait features a variety of brushstrokes, both thick and thin, resulting in blurred shapes and outlines while still providing a comprehensible illustration of an opera scene. On the other hand, `The Mechanic' by Fernand Léger (Figure~\ref{Figure:art_style_corr_lp} (c)), which depicts a working man in a futuristic modern industrial society, is classified as more Linear. This is due to the mechanic's distinctly tubular-shaped body and the intricate, sharply defined geometric shapes in the background. The Large Trademark with Eight Spotlights' by Edward Ruscha ((Figure~\ref{Figure:art_style_corr_lp} (d)) is also categorized as more Linear. This classification stems from its depiction of a logo in a long horizontal format, reminiscent of an opening credit scene in movies, with three-dimensional typography rendered in sharp linear perspective.

\begin{figure}[t]
  \centering
  \includegraphics[width=0.9\columnwidth]{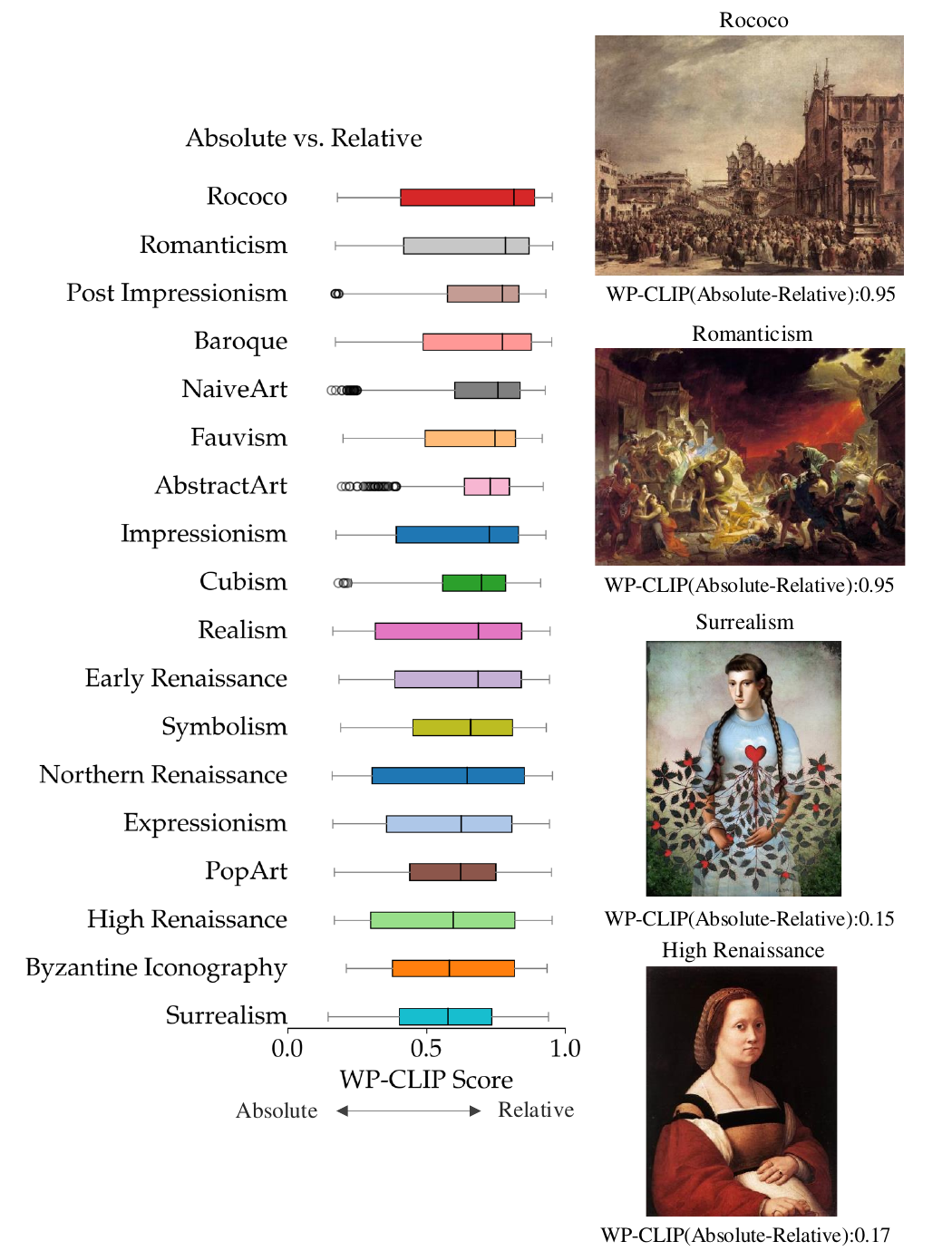}
  \caption{Predicted Absolute-Relative scores by WP-CLIP for 18 art movements in the Pandora-18k art dataset.}
  \label{Figure:art_style_corr_ar}
  \vspace{-0.15in}
\end{figure}

\begin{figure}[t!]
  \centering
  \includegraphics[width=0.6\columnwidth]{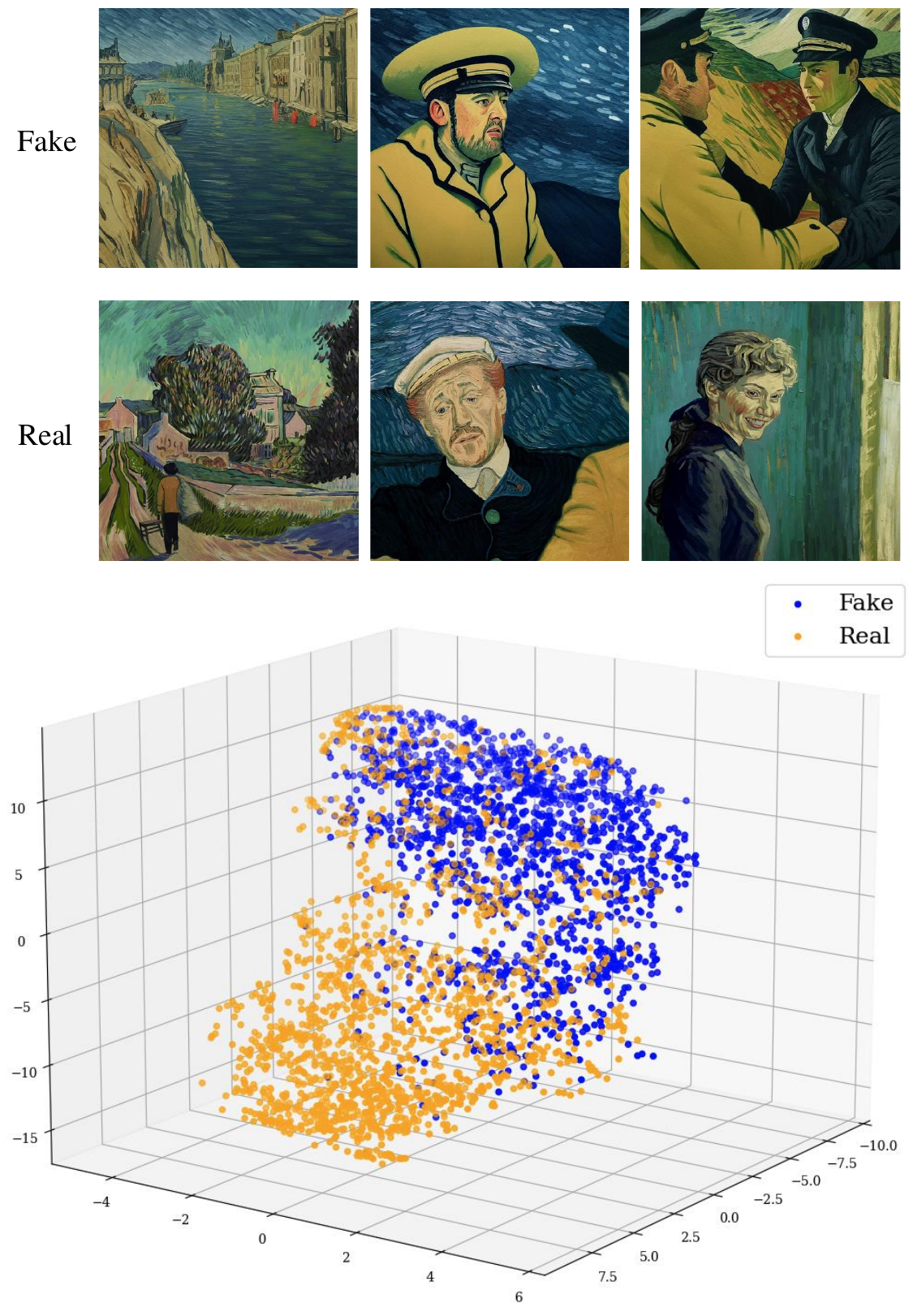}
  \vspace{-0.1in}
  \caption{The t-SNE projection shows how predicted scores may distinguish real and fake images, highlighting formal analysis's role in synthetic art detection."}
  \vspace{-0.2in}
  \label{Figure:fake_real}
\end{figure}

The Open Form principle extends beyond the artwork's edges, suggesting a sense of unending space beyond the field of view. In contrast, in the Closed Form principle, art is self-contained, concentrating the observer's attention on the content within its boundaries. The Open form principle is evident in paintings such as `Blue Courtyard, Arenys de Munt' by Santiago Rusiñol and `Boathouse in Argenteuil' by Gustave Caillebotte (Figure~\ref{Figure:art_style_corr_co} (a) and (b)). These works convey a strong sense of a specific place, enriched by the surrounding environment, which contributes to the feeling of continuity beyond the canvas. On the contrary, the Byzantine Icons art (Figure~\ref{Figure:art_style_corr_co} (c)) features religious figures focused on the subject within the image. Similarly, the painting `Saint Anthony Abbot and Michael the Archangel' by Filippo Lippi (Figure~\ref{Figure:art_style_corr_co} (d)) is a pair of panels from a triptych and employs linear perspective to depict full-body portraits of two significant figures, Saint Anthony the Abbot and Saint Michael, adopting a more Closed Form approach.

A composition that follows the Planar principle organizes elements into parallel planes with clear layers from foreground to background. In contrast, a composition that adheres more to the Recessional principle arranges objects and contrasting colors to depict spatial depth through diagonal lines and dynamic movement. Baroque art typically incorporates diagonal lines that recede into the picture plane, and our metric effectively identifies this characteristic in the painting shown in Figure~\ref{Figure:art_style_corr_pr} (a), classifying it as more Recessional. Similarly, The Woman on the Bank of the River by Pyotr Konchalovsky conveys a sense of depth, with the portrait subject in the foreground and the river in the distance, enhancing the feeling of depth in the background landscape. The Early Renaissance painting and Byzantine Icons art (Figure~\ref{Figure:art_style_corr_pr} (c) and (d)) do not depict depth in the same way; instead, they have elements arranged in layers, making them clearly more Planar. Our metric accurately categorizes paintings from Fauvism, Post-Impressionism, and Baroque as more Recessional, while High Renaissance, Early Renaissance, and Byzantine Iconography are identified as more Planar.

Paintings that adhere to the Unity principle feature less distinguishable figures and objects, appearing more blended together. In contrast, the Multiplicity principle emphasizes clear, distinct figures and separate elements. For example, `Village at Sunset' by Van Gogh (Figure~\ref{Figure:art_style_corr_mu}(a)) and the Rococo painting of fruit assortments (Figure~\ref{Figure:art_style_corr_mu}(b)) exhibit darker tones, causing colors to merge and figures to appear fused. On the other hand, in the pop art piece `The Defenders' by Rosalyn Drexler (Figure~\ref{Figure:art_style_corr_mu}(c)) and `Virgin Annunciate' by Fra Angelico (Figure~\ref{Figure:art_style_corr_mu}(d)), the figures are distinctly separated with clearly defined boundaries.

\begin{figure}[t!]
  \centering
  \includegraphics[width=0.85\columnwidth]{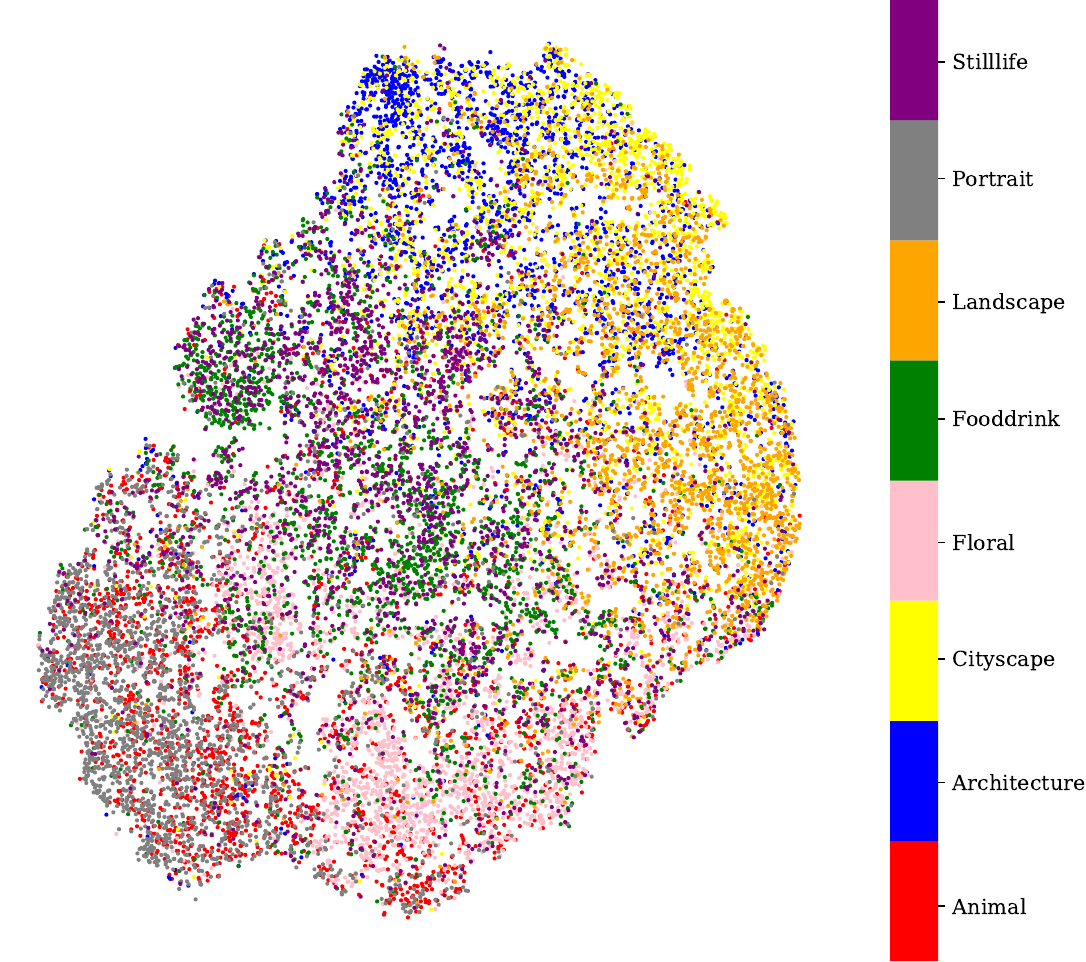}
  \vspace{-0.1in}
  \caption{This t-SNE projection illustrates how predicted scores provide meaningful features for distinguishing photography styles.}
  \vspace{-0.2in}
  \label{Figure:ava}
\end{figure}

In Absolute Clarity, figures and objects are brighter or illuminated, with well-defined edges, ensuring that forms are sharply distinguishable, whereas in Relative Clarity, boundaries are softer and less defined. As a result, Absolute Clarity ensures that every detail in the painting is clearly visible, while Relative Clarity focuses more on the overall composition rather than individual elements. This is evident in the Rococo and Romanticism paintings in Figure~\ref{Figure:art_style_corr_ar} (a) and (b), where the figures blend into a more cohesive composition, demonstrating Relative clarity. In contrast, the painting by Catrin Welz-Stein and the High Renaissance artworks feature distinctly defined figures, exhibiting a stronger sense of Absolute clarity.


\subsection{Synthetic Art Detection}\label{synthetic_art}

We present the results of our analysis using WP-CLIP’s predicted scores as features for synthetic art detection on the dataset from~\cite{bird2023ai}, shown in Figure~\ref{Figure:fake_real}. We project the five principles into three dimensions using t-SNE~\cite{van2008visualizing} to examine whether WP-CLIP’s predicted scores form distinct clusters for real and fake art images. In the figure, real and fake images appear visually similar, making them difficult to distinguish. However, WP-CLIP’s predicted scores seem to separate real and synthetic images into distinct clusters. This suggests that formal analysis based on Wölfflin's principles could be effective for synthetic art detection.

\subsection{Photography Style Recognition}\label{photo_styles}

Following the same approach as before in Section~\ref{synthetic_art}, we project the scores for photography styles in the AVA dataset~\cite{ava} using t-SNE. Figure~\ref{Figure:ava} shows that images of distinct photography styles tend to cluster around their respective centroids. 'Cityscape' and 'landscape' images are close together, with 'architecture' images nearby. Similarly, 'fooddrink,' 'floral,' and 'still life' styles form a neighboring group. Additionally, 'animal' images are positioned closer to 'portrait' styles. This experiment suggests that formal analysis via Wölfflin's principles provide a meaningful set of distinguishable features for photography styles.





\subsection{Art Generation Using WP-CLIP Guidance}

\begin{figure}[t!]
  \centering
  \includegraphics[width=0.95\columnwidth]{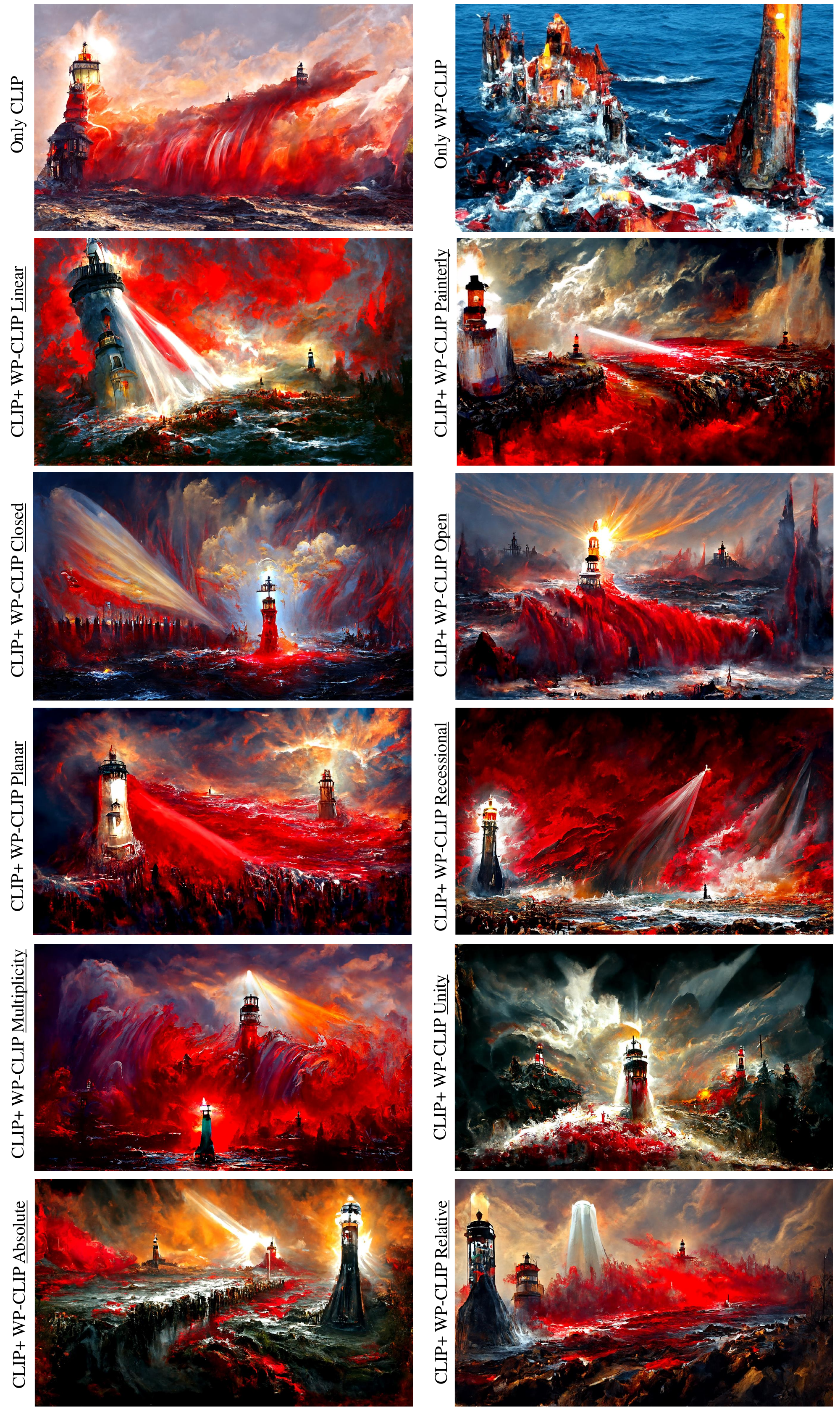}
  \vspace{-0.1in}
  \caption{Artistic images generated with Disco Diffusion, guided by CLIP and WP-CLIP. Prompt: `A beautiful painting of a singular lighthouse, shining its light across a tumultuous sea of blood by greg rutkowski and thomas kinkade. Trending on artstation.'}
  \label{Figure:diffusion}
  \vspace{-0.1in}
\end{figure}

We make use of Disco Diffusion~\cite{disco_diffusion}, a CLIP-guided diffusion based AI image generator capable of utilizing multiple CLIP models to create visually appealing art images. Initially, we experiment with standard CLIP guidance before incorporating WP-CLIP for diffusion control. As shown in Figure~\ref{Figure:diffusion}, our WP-CLIP model has forgotten the original CLIP model’s semantic knowledge, becoming more attuned to Wölfflin’s principles. To achieve a balance, we integrate both CLIP and WP-CLIP in the image generation process, allowing WP-CLIP to infuse the specified principle into the output. As demonstrated in Figure~\ref{Figure:diffusion}, this approach enables us to generate variations that align with the desired Wölfflin's principles in the resulting artwork.

\section{Conclusion}

We present WP-CLIP, a model that learns to predict contrasting Wölfflin's principles. Predicting artistic concepts based on the five Wölfflin's principles solely from visual elements is challenging due to their complex and nuanced nature. By fine-tuning a large pre-trained vision-language model, we leverage rich semantic features to learn scores for Wölfflin's principles. Through comprehensive experiments, we demonstrated its applicability in various visual art analyses, including assessing artistic style, art movement attribution, synthetic art detection, and photographic style recognition. Our findings highlight the relevance of formal analysis based on visual elements of art for these diverse tasks. Beyond analysis, WP-CLIP has the potential to guide the generation of art, making it a valuable tool as generative art becomes increasingly prevalent.

A recent study emphasized that Wölfflin's principles are not atemporal and must be expanded with additional concepts to effectively account for the evolving characteristics of generative art~\cite{deliege2025revisiting}. Complementary research has explored formal analysis by prompting LLMs with various types of concepts~\cite{kim2022proxy,tao2024does}. Alongside these advancements, our framework can incorporate additional concept pairs, broadening its applicability in future research.



{
    \small
    \bibliographystyle{ieeenat_fullname}
    \bibliography{main}

\begin{thebibliography}{39}
\providecommand{\natexlab}[1]{#1}
\providecommand{\url}[1]{\texttt{#1}}
\expandafter\ifx\csname urlstyle\endcsname\relax
  \providecommand{\doi}[1]{doi: #1}\else
  \providecommand{\doi}{doi: \begingroup \urlstyle{rm}\Url}\fi

\bibitem[Alembics and contributors(2021)]{disco_diffusion}
Alembics and contributors.
\newblock Disco diffusion, 2021.

\bibitem[Arnheim(1954)]{arnheim1954art}
Rudolf Arnheim.
\newblock \emph{Art and visual perception: A psychology of the creative eye}.
\newblock Univ of California Press, 1954.

\bibitem[Bianco et~al.(2019)Bianco, Mazzini, Napoletano, and Schettini]{bianco2019multitask}
Simone Bianco, Davide Mazzini, Paolo Napoletano, and Raimondo Schettini.
\newblock Multitask painting categorization by deep multibranch neural network.
\newblock \emph{Expert Systems with Applications}, 135:\penalty0 90--101, 2019.

\bibitem[Bin et~al.(2024)Bin, Shi, Ding, Hu, Wang, Yang, Ng, and Shen]{bin2024gallerygpt}
Yi Bin, Wenhao Shi, Yujuan Ding, Zhiqiang Hu, Zheng Wang, Yang Yang, See-Kiong Ng, and Heng~Tao Shen.
\newblock Gallerygpt: Analyzing paintings with large multimodal models.
\newblock In \emph{ACM International Conference on Multimedia}, pages 7734--7743, 2024.

\bibitem[Bird et~al.(2023)Bird, Barnes, and Lotfi]{bird2023ai}
Jordan~J Bird, Chloe~M Barnes, and Ahmad Lotfi.
\newblock Ai generated art: Latent diffusion-based style and detection.
\newblock In \emph{UK Workshop on Computational Intelligence}, pages 157--169. Springer, 2023.

\bibitem[Castellano and Vessio(2021)]{castellano2021deep}
Giovanna Castellano and Gennaro Vessio.
\newblock Deep convolutional embedding for digitized painting clustering.
\newblock In \emph{International Conference on Pattern Recognition}, pages 2708--2715. IEEE, 2021.

\bibitem[Cetinic et~al.(2018)Cetinic, Lipic, and Grgic]{cetinic2018fine}
Eva Cetinic, Tomislav Lipic, and Sonja Grgic.
\newblock Fine-tuning convolutional neural networks for fine art classification.
\newblock \emph{Expert Systems with Applications}, 114:\penalty0 107--118, 2018.

\bibitem[Cetinic et~al.(2020)Cetinic, Lipic, and Grgic]{cetinic2020learning}
Eva Cetinic, Tomislav Lipic, and Sonja Grgic.
\newblock Learning the principles of art history with convolutional neural networks.
\newblock \emph{Pattern Recognition Letters}, 129:\penalty0 56--62, 2020.

\bibitem[Chow et~al.(2025)Chow, Mao, Li, Seita, Guizilini, and Wang]{chow2025physbench}
Wei Chow, Jiageng Mao, Boyi Li, Daniel Seita, Vitor~Campagnolo Guizilini, and Yue Wang.
\newblock Physbench: Benchmarking and enhancing vision-language models for physical world understanding.
\newblock In \emph{International Conference on Learning Representations}, 2025.

\bibitem[Conde and Turgutlu(2021)]{conde2021clip}
Marcos~V Conde and Kerem Turgutlu.
\newblock Clip-art: Contrastive pre-training for fine-grained art classification.
\newblock In \emph{IEEE/CVF Conference on Computer Vision and Pattern Recognition}, pages 3956--3960, 2021.

\bibitem[Deliege et~al.(2025)Deliege, Dondero, and D’Armenio]{deliege2025revisiting}
Adrien Deliege, Maria~Giulia Dondero, and Enzo D’Armenio.
\newblock Revisiting w{\"o}lfflin in the age of ai: A study of classical and baroque composition in generative models.
\newblock \emph{Journal of Imaging}, 11\penalty0 (5):\penalty0 128, 2025.

\bibitem[Elgammal et~al.(2017)Elgammal, Liu, Elhoseiny, and Mazzone]{elgammal2017can}
Ahmed Elgammal, Bingchen Liu, Mohamed Elhoseiny, and Marian Mazzone.
\newblock Can: Creative adversarial networks generating “art” by learning about styles and deviating from style norms.
\newblock In \emph{International Conference on Computational Creativity}, 2017.

\bibitem[Elgammal et~al.(2018)Elgammal, Liu, Kim, Elhoseiny, and Mazzone]{elgammal2018shape}
Ahmed Elgammal, Bingchen Liu, Diana Kim, Mohamed Elhoseiny, and Marian Mazzone.
\newblock The shape of art history in the eyes of the machine.
\newblock In \emph{AAAI Conference on Artificial Intelligence}, 2018.

\bibitem[Florea et~al.(2017)Florea, Toca, and Gieseke]{florea2017artistic}
Corneliu Florea, Cosmin Toca, and Fabian Gieseke.
\newblock Artistic movement recognition by boosted fusion of color structure and topographic description.
\newblock In \emph{IEEE Winter Conference on Applications of Computer Vision (WACV)}, pages 569--577. IEEE, 2017.

\bibitem[Foka(2024)]{foka2024framework}
Amalia Foka.
\newblock A framework for critical evaluation of text-to-image models: Integrating art historical analysis, artistic exploration, and critical prompt engineering.
\newblock \emph{arXiv preprint arXiv:2412.12774}, 2024.

\bibitem[Frank et~al.(2019)Frank, Preble, and Preble]{frank2019prebles}
Patrick Frank, Duane Preble, and Sarah Preble.
\newblock \emph{Prebles' Artforms}.
\newblock Pearson, 2019.

\bibitem[Gairola et~al.(2020)Gairola, Shah, and Narayanan]{gairola2020unsupervised}
Siddhartha Gairola, Rajvi Shah, and PJ Narayanan.
\newblock Unsupervised image style embeddings for retrieval and recognition tasks.
\newblock In \emph{IEEE/CVF Winter Conference on Applications of Computer Vision}, pages 3281--3289, 2020.

\bibitem[Garcia et~al.(2019)Garcia, Renoust, and Nakashima]{garcia2019context}
Noa Garcia, Benjamin Renoust, and Yuta Nakashima.
\newblock Context-aware embeddings for automatic art analysis.
\newblock In \emph{International Conference on Multimedia Retrieval}, pages 25--33, 2019.

\bibitem[Gombrich(1960)]{gombrich1960study}
Ernst~Hans Gombrich.
\newblock \emph{A Study in the Psychology of Pictorial Representation}.
\newblock Pantheon Books, 1960.

\bibitem[Google(2025)]{google_gemini25pro_2025}
Google.
\newblock Gemini 2.5 pro: Our most intelligent ai model.
\newblock \url{https://blog.google/technology/google-deepmind/gemini-model-thinking-updates-march-2025/}, 2025.
\newblock Accessed: 2025-04-21.

\bibitem[Hatt(2006)]{hatt2006art}
Michael Hatt.
\newblock \emph{Art History: A Critical Introduction to Its Methods}.
\newblock Manchester University Press, 2006.

\bibitem[Jha et~al.(2021)Jha, Chang, and Elhoseiny]{djwaga}
Divyansh Jha, Hanna~H. Chang, and Mohamed Elhoseiny.
\newblock Wölfflin’s affective generative analysis of visual art.
\newblock \emph{The International Conference on Computational Creativity (ICCC)}, 2021.

\bibitem[Karras et~al.(2020)Karras, Laine, Aittala, Hellsten, Lehtinen, and Aila]{karras2020analyzing}
Tero Karras, Samuli Laine, Miika Aittala, Janne Hellsten, Jaakko Lehtinen, and Timo Aila.
\newblock Analyzing and improving the image quality of stylegan.
\newblock In \emph{IEEE/CVF conference on computer vision and pattern recognition}, pages 8110--8119, 2020.

\bibitem[Khan et~al.(2024)Khan, Kim, Jha, Mohamed, Chang, Elgammal, Elliott, and Elhoseiny]{khan2024ai}
Faizan~Farooq Khan, Diana Kim, Divyansh Jha, Youssef Mohamed, Hanna~H Chang, Ahmed Elgammal, Luba Elliott, and Mohamed Elhoseiny.
\newblock Ai art neural constellation: Revealing the collective and contrastive state of ai-generated and human art.
\newblock In \emph{IEEE/CVF Conference on Computer Vision and Pattern Recognition Workshop}, pages 7470--7478, 2024.

\bibitem[Kim et~al.(2022)Kim, Elgammal, and Mazzone]{kim2022proxy}
Diana Kim, Ahmed Elgammal, and Marian Mazzone.
\newblock Proxy learning of visual concepts of fine art paintings from styles through language models.
\newblock In \emph{AAAI Conference on Artificial Intelligence}, pages 4513--4522, 2022.

\bibitem[Liu et~al.(2021)Liu, Yang, Agaian, and Yuan]{liu2021novel}
Shao Liu, Jiaqi Yang, Sos~S Agaian, and Changhe Yuan.
\newblock Novel features for art movement classification of portrait paintings.
\newblock \emph{Image and Vision Computing}, 108:\penalty0 104121, 2021.

\bibitem[Lu et~al.(2024)Lu, Bansal, Xia, Liu, Li, Hajishirzi, Cheng, Chang, Galley, and Gao]{lu2024mathvista}
Pan Lu, Hritik Bansal, Tony Xia, Jiacheng Liu, Chunyuan Li, Hannaneh Hajishirzi, Hao Cheng, Kai-Wei Chang, Michel Galley, and Jianfeng Gao.
\newblock Mathvista: Evaluating mathematical reasoning of foundation models in visual contexts.
\newblock In \emph{International Conference on Learning Representations}, 2024.

\bibitem[Murray et~al.(2012)Murray, Marchesotti, and Perronnin]{ava}
Naila Murray, Luca Marchesotti, and Florent Perronnin.
\newblock Ava: A large-scale database for aesthetic visual analysis.
\newblock In \emph{IEEE/CVF Conference on Computer Vision and Pattern Recognition}, pages 2408--2415, 2012.

\bibitem[Radford et~al.(2021)Radford, Kim, Hallacy, Ramesh, Goh, Agarwal, Sastry, Askell, Mishkin, Clark, et~al.]{radford2021learning}
Alec Radford, Jong~Wook Kim, Chris Hallacy, Aditya Ramesh, Gabriel Goh, Sandhini Agarwal, Girish Sastry, Amanda Askell, Pamela Mishkin, Jack Clark, et~al.
\newblock Learning transferable visual models from natural language supervision.
\newblock In \emph{International Conference on Machine Learning}, pages 8748--8763, 2021.

\bibitem[Sigaki et~al.(2018)Sigaki, Perc, and Ribeiro]{sigaki2018history}
Higor~YD Sigaki, Matja{\v{z}} Perc, and Haroldo~V Ribeiro.
\newblock History of art paintings through the lens of entropy and complexity.
\newblock \emph{Proceedings of the National Academy of Sciences}, 115\penalty0 (37):\penalty0 E8585--E8594, 2018.

\bibitem[Taesiri et~al.(2025)Taesiri, Ghildyal, Zadtootaghaj, Barman, and Bezemer]{taesiri2025videogameqa}
Mohammad~Reza Taesiri, Abhijay Ghildyal, Saman Zadtootaghaj, Nabajeet Barman, and Cor-Paul Bezemer.
\newblock Videogameqa-bench: Evaluating vision-language models for video game quality assurance.
\newblock \emph{arXiv:2505.15952}, 2025.

\bibitem[Tao and Xie(2024)]{tao2024does}
Muzi Tao and Saining Xie.
\newblock What does a visual formal analysis of the world's 500 most famous paintings tell us about multimodal llms?
\newblock In \emph{Tiny Papers Track at Internation Conference on Learning Representations}, 2024.

\bibitem[Van~der Maaten and Hinton(2008)]{van2008visualizing}
Laurens Van~der Maaten and Geoffrey Hinton.
\newblock Visualizing data using t-sne.
\newblock \emph{Journal of Machine Learning Research}, 9\penalty0 (11), 2008.

\bibitem[Wang et~al.(2023)Wang, Chan, and Loy]{wang2022exploring}
Jianyi Wang, Kelvin~CK Chan, and Chen~Change Loy.
\newblock Exploring clip for assessing the look and feel of images.
\newblock In \emph{AAAI Conference on Artificial Intelligence}, 2023.

\bibitem[W{\"o}lfflin and Hottinger(1950)]{wolfflin1950principles}
Heinrich W{\"o}lfflin and Marie~Donald Hottinger.
\newblock \emph{Principles of art history: The problem of the development of style in later art}.
\newblock Dover New York, 1950.

\bibitem[Wright and Ommer(2022)]{wright2022artfid}
Matthias Wright and Bj{\"o}rn Ommer.
\newblock Artfid: Quantitative evaluation of neural style transfer.
\newblock In \emph{DAGM German Conference on Pattern Recognition}, pages 560--576. Springer, 2022.

\bibitem[Yue et~al.(2024)Yue, Ni, Zhang, Zheng, Liu, Zhang, Stevens, Jiang, Ren, Sun, et~al.]{yue2024mmmu}
Xiang Yue, Yuansheng Ni, Kai Zhang, Tianyu Zheng, Ruoqi Liu, Ge Zhang, Samuel Stevens, Dongfu Jiang, Weiming Ren, Yuxuan Sun, et~al.
\newblock Mmmu: A massive multi-discipline multimodal understanding and reasoning benchmark for expert agi.
\newblock In \emph{IEEE/CVF Conference on Computer Vision and Pattern Recognition}, pages 9556--9567, 2024.

\bibitem[Zhang et~al.(2022)Zhang, Xiao, Zhou, Xia, Xie, and Liu]{zhang2022analysis}
Tingting Zhang, Shuang Xiao, Wei Zhou, Ling Xia, Jinwei Xie, and Xiaofeng Liu.
\newblock Analysis of painting complexity based on feature extraction.
\newblock In \emph{International Conference on Advances in Computer Technology, Information Science and Communications (CTISC)}, pages 1--5. IEEE, 2022.

\bibitem[Zou et~al.(2014)Zou, Cao, Li, Huang, and Wang]{zou2014chronological}
Qin Zou, Yu Cao, Qingquan Li, Chuanhe Huang, and Song Wang.
\newblock Chronological classification of ancient paintings using appearance and shape features.
\newblock \emph{Pattern Recognition Letters}, 49:\penalty0 146--154, 2014.

\end{thebibliography}
}

\end{document}